%% file: autoenc_top_acm.tex
\documentclass[9pt,sigconf,nonacm,pbalance=true,natbib=false]{acmart}

\setcopyright{none}
\copyrightyear{2025}
\acmYear{2025}
\acmDOI{XXXXXXX.XXXXXXX}
\acmConference[XXXXXX] {Temp}{2025} {XXXXX}
\acmISBN{978-1-4503-XXXX-X/2025/11}

\usepackage{algorithmic}
\usepackage{graphicx}
\usepackage{textcomp}
\usepackage{makecell}
\usepackage{xcolor}
\usepackage[table]{xcolor}
\usepackage{tcolorbox}
\usepackage{balance}
\usepackage{array}
\usepackage{nicematrix}

\usepackage{listings}
\definecolor{headerblue}{HTML}{2C508A}
\definecolor{headerwhite}{HTML}{FFFFFF}

\newcolumntype{I}{!{\vrule}}
\newcommand{\ThickXhline}{1.5}

\RequirePackage[
  datamodel=acmdatamodel,
  style=acmnumeric,
  ]{biblatex}

\begin{document}

\title{Mitigating hallucinations and omissions in LLMs for invertible problems: An application to hardware logic design automation}



\author{Andrew S. Cassidy, Guillaume Garreau, Jay Sivagnaname, Mike Grassi, Bernard Brezzo, \\ John V. Arthur, Dharmendra S. Modha}
\affiliation{
  \institution{IBM Research}
  \city{ }
  \country{ }
}

\input{abstract.tex}

\maketitle


\input{introduction.tex}

\input{background.tex}

\input{approach.tex}

\input{experiment1_RTR.tex}

\input{results.tex}

\input{discussion.tex}

\input{conclusion.tex}



\printbibliography


\end{document}

%% file: abstract.tex
\begin{abstract}
We show for invertible problems that transform data from a source domain
(for example, Logic Condition Tables (LCTs))
to a destination domain (for example, Hardware Description Language (HDL) code),
an approach of using Large Language Models (LLMs) as a lossless encoder
from source to destination followed by as a lossless decoder back to the source,
comparable to lossless compression in information theory,
can mitigate most of the LLM drawbacks of hallucinations and omissions.
Specifically, using LCTs as inputs, we generate the full HDL for a
two-dimensional network-on-chip router (13 units, 1500-2000 lines of code) using seven different LLMs,
reconstruct the LCTs from the auto-generated HDL, and
compare the original and reconstructed LCTs.
This approach yields significant productivity improvements,
not only confirming correctly generated LLM logic and detecting incorrectly generated LLM logic
but also assisting developers in finding design specification errors.
\end{abstract}

%% file: introduction.tex
\section{Introduction} \label{sec:introduction}

Large Language Models (LLMs) suffer from the fundamental challenges of false positives (hallucinations) and 
false negatives (omitted information).  Recent research shows that hallucination is inevitable 
\cite{xu2025hallucinationinevitableinnatelimitation}
and LLMs have difficulty with omissions
\cite{fu2025absencebenchlanguagemodelscant}.
We demonstrate that it is possible to mitigate these challenges for invertible problems 
consisting of essentially translating the same information into different semantic contexts.
Without directly combating intrinsic false positives and false negatives inherent in the LLMs,
we propose a direction whereby it is possible to spot and correct these extrinsically. 
Eschewing the fundamental scientific and philosophical questions about whether LLMs
understand semantic meaning or operate syntactically on form \cite{bender2020climbing},
we adopt an utilitarian and engineering approach of focusing on how to direct LLMs to produce
correct, complete, and concise output for invertible problems.

Comparable to lossless compression in information theory, we propose using LLMs as an encoder 
(from a source context into a destination) followed by as a decoder (from the destination context back into the source), 
and then verifying that this results in an identity map. 
The input itself becomes the ground truth for axiomatically asserting correctness \cite{hoare1969axiomatic}.
Note that while this approach mitigates hallucinations and omissions, it does not mathematically eliminate them
because either the encoder phase, the decoder phase, or both might introduce self-cancelling hallucinations producing a match.
Therefore, formal verification is still necessary.
Many classes of such practical problems exist, 
for example, translating COBOL into Java/C++, translating English into Spanish, and 
translating JSON into XML/YAML/CSV formats. 
This approach is grounded in recent results that language models are injective and hence invertible
\cite{nikolaou2025languagemodelsinjectiveinvertible}
and is inspired by well studied autoencoders for lossy compression \cite{hecht1995replicator}.

As an illustrative example, we demonstrate that this approach can correctly translate Logic Condition Tables (LCTs) 
to Hardware Description Language (HDL) code, 
while detecting both LLM encoding/decoding errors and assisting 
developers in finding inherent design specification errors, thus significantly increasing designer productivity.

\begin{figure}[tbp]
\centerline{\includegraphics[width=\columnwidth]{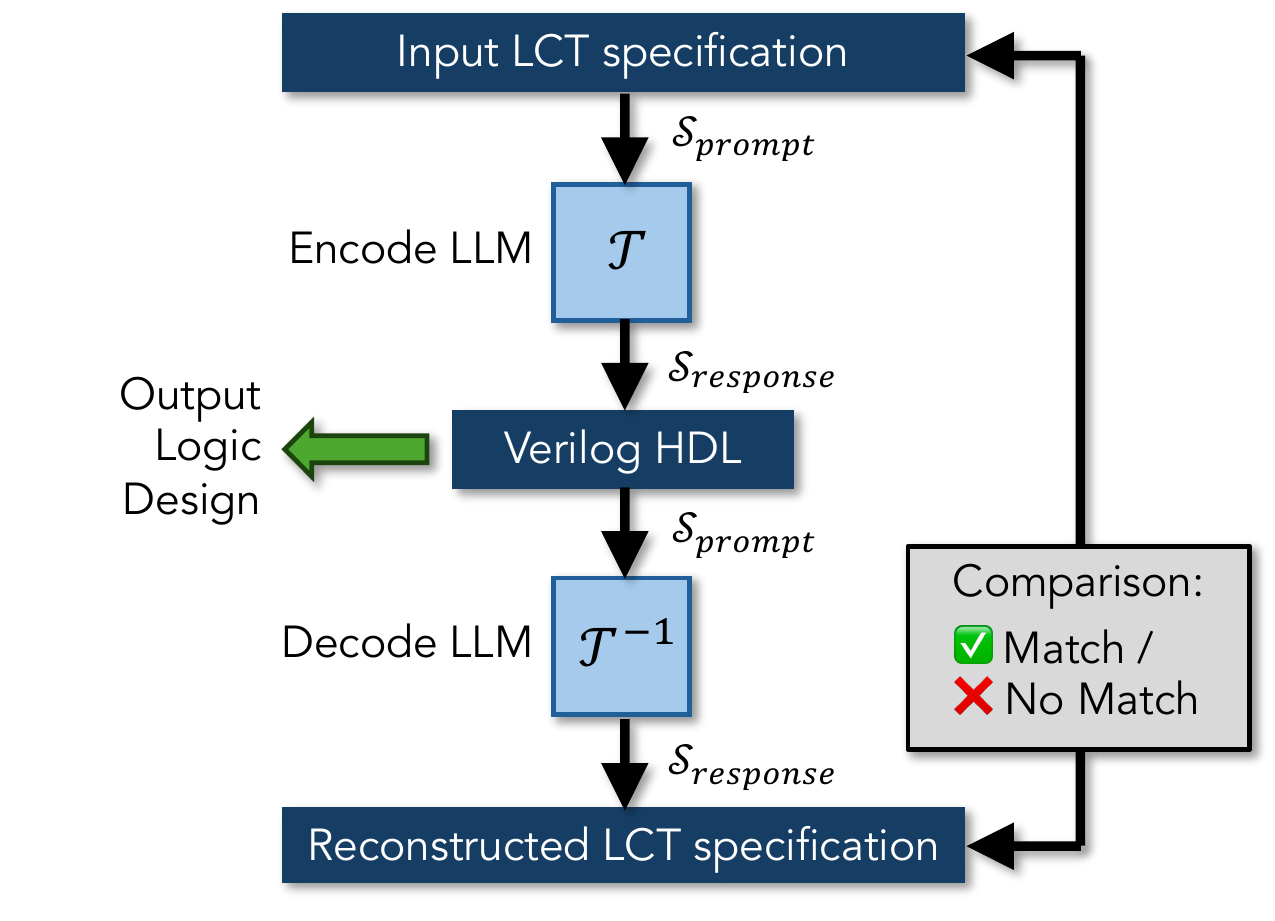}}
\caption{View of an LLM as an invertible transform $\mathcal{T}$
from an input token sequence $\mathcal{S}_{prompt}$
to an output token sequence $\mathcal{S}_{response}$.
The input prompt sequence is a specification.
The output sequence contains the completed HDL code.
By reapplying the LLM transform $\mathcal{T}^{-1}$ in the reverse direction,
the input specification can be reconstructed from the output HDL code and checked
versus the original specification.
}
\label{fig:LLM_autoenc}
\end{figure}


Precise specification using LCTs
ensures all design information is completely and correctly defined.
An LCT succinctly specifies system input conditions and corresponding
output results in table columns for each case of values in table rows
(see Section \ref{sec:approach}).
The closed loop solution using LLMs as an invertible transform
to verify the implementation is shown in Fig. \ref{fig:LLM_autoenc}.
An LLM performs a transform $\mathcal{T}$
from an input token sequence $\mathcal{S}_{prompt}$
to an output token sequence $\mathcal{S}_{response}$.
Mathematically, the transform is defined as:
$\mathcal{S}_{response} = \mathcal{T} ( \mathcal{S}_{prompt} )$
where the transform input and output are token sequences:
$\mathcal{S}_{prompt} = \{s_0, s_1, s_2, ..., s_{n-1} \}$ and
$\mathcal{S}_{response} = \{s_0, s_1, s_2, ..., s_{m-1} \}$ respectively.
The LLM transform $\mathcal{T}$ can be applied in the forward or encode direction to generate
Verilog HDL from a specification.
Or the LLM transform $\mathcal{T}^{-1}$ can be applied in the inverse or decode direction
to reconstruct the original LCT specification from the Verilog HDL.
Applying these encoder / decoder transforms forms an autoencoder.

Using the autoencoder as a verification method, the original input LCT specification and the reconstructed
output LCT specification are compared.  If the input and output LCTs don't match, one of three cases is true:
\begin{enumerate}
\item an error in the encode/forward transform (LLM error),
\item error in the decode/inverse transform (LLM error),
\item or inadequate original specification (designer error).
\end{enumerate}
As a result, our approach is a verification method
primarily for checking the LLM's work and only secondarily the LCT specification.
Note that our approach's ability to check the input LCT specification itself is limited.
If the designer defines the wrong function with a complete, well-formed input LCT specification,
our approach will correctly implement the wrong function.
Thus, functional verification of the input specification and overall design is still necessary,
as well as to cover the case of self-cancelling hallucinations.

%% file: background.tex
\section{Background} \label{sec:background}


Boolean Truth Tables \cite{wittgenstein1922,post1921introduction}
form the basis of digital logic design.
Programmable digital computing circuits,
such as EDVAC \cite{vonNeumann:1945:EDVAC} and ENIAC \cite{hartree1946eniac},
followed closely after.
As succeeding decades brought innovations in programming these computing machines,
Decision Tables \cite{pollack1963analysis,pooch1974translation,codasyl1982modern,vanthienen1997decision}
were developed as a means for specifying programatic thinking.
They concisely specify the actions to take for any given input condition to the system in a table format.
Boolean Truth Tables and Decision Tables are the foundation of the LCTs 
described in Section \ref{sec:approach}.
LCTs were first applied to auditing the logic of a 22B transistor
Neural Inference Processor \cite{np_science,np_isscc} prior to tapeout.
Used as a formal verification methodology, over 160 LCTs were manually 
constructed from the Verilog source code, one per process statement on selected logic units.
The tables, representing the underlying logic, were then checked for correctness and completeness.
As a result, the processor was successful in first pass silicon.

%
A wide variety of approaches containing natural language descriptions of logic designs,
have been developed to test LLM approaches for automated logic generation.
Examples include:
model fine tuning \cite{thakur2023verigenlargelanguagemodel},
data augmentation \cite{gao2024autovcodersystematicframeworkautomated,min2025improving,calzada2025verilogdblargesthighestqualitydataset,Roberts2025},
instruction tuning \cite{zhao2025codevempoweringllmshdl},
feedback-directed refinement \cite{wei2025vericoderenhancingllmbasedrtl,blocklove2025automaticallyimproving},
and extracting logic equations from natural language \cite{roy2025veritasdeterministicverilogcode}.
%
%
%
%

%% file: approach.tex
\section{Approach} \label{sec:approach}

\subsection{Logic Condition Tables (LCTs)} \label{subsec:lcts}

\input{table_lct_comb.tex}

\input{table_lct_seq.tex}

\input{table_fsm_lct.tex}

\input{table_hier_conn.tex}

LCTs are an efficient representation of logic design blocks,
including both combinational and sequential circuits.
Derived from Decision Tables and their cousins, the Boolean Truth Table,
examples of the basic LCT structure are shown in
Table \ref{tab:lct_comb} (combinational 4-input MUX),
Table \ref{tab:lct_seq} (registered 2-input MUX), and
Table \ref{tab:fsm_lct} (4-state Finite State Machine).
Table columns contain
    the set of input conditions to the design and 
    the set of output results/actions produced by the design.
Table rows contain
    the set of cases, which define the result to produce (or action to take),
    given the specific set of input conditions.
The first column, ``Case,'' is not required, it is just shown for illustration.
The last column, ``Comments,'' is optional, however, it is useful for human table designers
(and potentially for the LLM as well---but that hypothesis is untested in this paper).
Observe that neither conditions nor outputs are required to be binary valued.
Condition headers may be a signal name, or may be an actual logical or arithmetic condition
(for example: $A \& \mathtt{\sim}B$ or $C \leq 10$).
The ``don't care'' symbol is ``X.''
In most logic designs there are many ``don't care'' cases, leading to a significant compression of rows
relative to the possible number of cases.
The table rows display a natural hierarchy of nested `if' and/or `case' statements.

Table \ref{tab:lct_seq} defines a simple registered 2-input multiplexor, with a data valid signal and
backpressure (ready) signal.
In the backpressure case, the $valid\_out$ and $data\_out$ registers should hold their value
(do not update), denoted by entering the output signal name into the corresponding table cell.
Table \ref{tab:fsm_lct} defines an example four state Finite State Machine (FSM),
where each state has two possible outgoing transitions, selected by input conditions 0 and 1.

\input{table_models.tex}


LCTs are particularly well suited for specifying the parallelism of logic circuits, as many output
signals can be driven by the same input condition set.  In addition, logic circuits can be
partitioned into multiple LCTs, subdividing the complexity, similar to parallel process blocks in HDL.
As the LCT itself does not specify clocking, we specify in the prompt whether each LCT is
``clocked'' or ``combinational''
to direct the LLM to implement a clocked process (sequential) circuit or unclocked (combinational) circuit.
The timing semantics are the same as standard logic design.
Clocked LCTs are state holding and retain their state from the prior clock cycle. 
Unclocked LCTs do not retain state from the prior clock cycle.

One significant advantage of the LCTs is that they compel human logic designers into a
disciplined and structured design methodology.
In constructing the LCT, human designers must think through both the conditions and the results
of each case.  All condition cases should be enumerated.
Checks, transforms, and optimizations can be applied to LCTs, however, full treatment of these subjects
is outside the scope of this paper.

An example connectivity table definition for hierarchical design is shown in
Table \ref{tab:hier_conn}.
A simple definition, similar to a standard HDL port map,
it defines the connectivity for all ports of a module.
Net context defines whether a signal interconnects to other modules or to the external ports
(up a level of hierarchy).
Connectivity tables are not the focus of this paper, however, they also proved to be an effective
specification for LLMs.

\subsection{Closed Loop Design Flow} \label{subsec:close_loop}

The closed loop design flow follows Fig. \ref{fig:LLM_autoenc}.
The automation of steps 2, 3, and 4 drive developer productivity improvement.
\begin{enumerate}
\item (Human) designer specifies the logic as an LCT.
\item Apply the LLM transform $\mathcal{T}$ to the LCT to generate the Verilog HDL.
\item Apply the LLM inverse transform $\mathcal{T}^{-1}$ to the generated Verilog HDL
    to reconstruct the original LCT specification.
\item Compare the original LCT specification with the reconstructed LCT specification.
\item Fix any errors (and repeat from 2. if necessary). 
\end{enumerate}
%


\input{fig_rtr.tex}



\subsection{Forward Transform Prompt} \label{subsec:fwd_prompt}
For the forward transform, the prompt is simple, using the LCT as the specification together with the Verilog port map.
The prompt consisted of the following:
\begin{enumerate}
\item Specify clocked or combinational Verilog.
\item Specify the LCT columns (\# input condition columns and \# output result columns).
\item List the LCT in CSV format.
\item Define the Verilog port map, including the module input and output signals.
\end{enumerate}

\subsection{Inverse Transform Prompt} \label{subsec:inv_prompt}
For the inverse transform, the prompt is more involved, as LCTs are unknown to LLMs and thus
LLMs have not been trained to generate LCTs.
The prompt consisted of the following:
\begin{enumerate}
\item Short natural language definition of an LCT.
\item List a full example of a Verilog HDL (MUX2 design).
\item List a full example of the corresponding LCT (MUX2 design).
\item List the Verilog to evaluate.
\item Define the input condition and output result column headers for the LCT to be reconstructed.
\end{enumerate}

\subsection{LLM Models} \label{subsec:ai_llm_models}

Table \ref{tab:models} summarizes the LLM Models used in this paper.
All models are publicly available and used ``off-the-shelf,''
without retraining, fine tuning, augmentation, or any other modification.

%% file: table_lct_comb.tex

\begin{table}[tbp]
\caption{Example combinational LCT: 4-input MUX. Table columns contain input conditions (blue) and output results (green).
	Table rows contain cases. 
}
\begin{center}
\small

\begin{NiceTabular}{|>{\centering} p{1.2cm}|>{\centering} p{1.2cm}|>{\centering} p{1.2cm}|>{\centering} p{1.2cm}|l|}
\hline

\cellcolor{gray!25}{Case}	& \multicolumn{2}{c|}{\cellcolor{cyan!25}{Inputs}}   &   \cellcolor{green!25}{Outputs} & \cellcolor{gray!25}{Comments}	\\
\hline
\cellcolor{gray!25}{ } &	\cellcolor{cyan!25}{enable} &	\cellcolor{cyan!25}{select} &
	\cellcolor{green!25}{data\_out} &	\cellcolor{gray!25}{ } \\
\Xhline{\ThickXhline pt}

0 &		0 &		X &		0 &		Disabled \\
\hline
1 &		1 &		0 &		data0 &		output $1^{st}$ data \\
\hline
2 &		1 &		1 &		data1 &		output $2^{nd}$ data \\
\hline
3 &		1 &		2 &		data2 &		output $3^{rd}$ data \\
\hline
4 &		1 &		3 &		data3 &		output $4^{th}$ data \\

\hline
\end{NiceTabular}
\label{tab:lct_comb}
\end{center}
\end{table}

%% file: table_lct_seq.tex

\begin{table}[tbp]
\caption{Example sequential LCT: registered 2-input MUX with data valid and backpressure (ready).
}
\begin{center}
\small

\begin{NiceTabular}{|c|c|c|c|c|c|l|}
\hline

\multicolumn{4}{|c|}{\cellcolor{cyan!25}{Inputs}}   &   \multicolumn{2}{c|}{\cellcolor{green!25}{Outputs}} & \cellcolor{gray!25}{Comments}	\\
\hline

\cellcolor{cyan!25}{rst\_n} &	\cellcolor{cyan!25}{ready} &	\cellcolor{cyan!25}{valid\_in} &	\cellcolor{cyan!25}{select} &
\cellcolor{green!25}{valid\_out} &	\cellcolor{green!25}{data\_out} &	\cellcolor{gray!25}{ } \\
\Xhline{\ThickXhline pt}

0 &		X &		X &		X &		0 &		0 &		Reset \\
\hline
1 &		0 &		X &		X &		valid\_out &	data\_out &		Backpress \\
\hline
1 &		1 &		0 &		X &		0 &		0 &		No input \\
\hline
1 &		1 &		1 &		0 &		1 &		data0 &		Select $0$ \\
\hline
1 &		1 &		1 &		1 &		1 &		data1 &		Select $1$ \\

\hline
\end{NiceTabular}
\label{tab:lct_seq}
\end{center}
\end{table}

%% file: table_fsm_lct.tex

\begin{table}[tbp]
\caption{Example Finite State Machine LCT with four states,
two outgoing transitions per state---selected by cond0 and cond1, and three output results.}
\begin{center}
\small

\begin{NiceTabular}{|c|c|c|c|c|c|c|c|}
\hline

\multicolumn{4}{|c|}{\cellcolor{cyan!25}{Inputs}}   &   \multicolumn{4}{c|}{\cellcolor{green!25}{Outputs}} \\
\hline

\cellcolor{cyan!25}{rst\_n} &	\cellcolor{cyan!25}{state} &	\cellcolor{cyan!25}{cond0} &	\cellcolor{cyan!25}{cond1} &	\cellcolor{green!25}{next\_state} &	\cellcolor{green!25}{out0} &	\cellcolor{green!25}{out1} &	\cellcolor{green!25}{out2} \\
\Xhline{\ThickXhline pt}

0 &		X &		X &		X &		0 &		0 &		0 &		0 \\
\hline
1 &		X &		0 &		0 &		state &		out0 &		out1 &		out2 \\
\hline
1 &		0 &		1 &		0 &		0 &		1 &		0 &		0 \\
\hline
1 &		0 &		0 &		1 &		2 &		1 &		0 &		0 \\
\hline
1 &		1 &		1 &		0 &		2 &		0 &		0 &		1 \\
\hline
1 &		1 &		0 &		1 &		3 &		0 &		0 &		1 \\
\hline
1 &		2 &		1 &		0 &		3 &		0 &		1 &		1 \\
\hline
1 &		2 &		0 &		1 &		0 &		0 &		1 &		1 \\
\hline
1 &		3 &		1 &		0 &		1 &		1 &		1 &		0 \\
\hline
1 &		3 &		0 &		1 &		1 &		1 &		1 &		0 \\

\hline
\end{NiceTabular}
\label{tab:fsm_lct}
\end{center}
\end{table}

%% file: table_hier_conn.tex

\begin{table}[tbp]
\caption{Hierarchical Connectivity Table: The lower row is repeated for all ports in a module instantiation.}
\begin{center}
\small

\begin{tabular}{|p{1.2cm}|p{1.2cm}|p{1.8cm}|p{1.2cm}|p{1.4cm}|}
\hline

Port Direction	& Port Name	& Net Name			& Size (bits) & Net Context \\
\Xhline{\ThickXhline pt}

$\{$input, output$\}$ & module signal & interconnection signal & \{\# bits\}  & \{internal, external\} \\


\hline
\end{tabular}
\label{tab:hier_conn}
\end{center}
\end{table}

%% file: table_models.tex

\begin{table}[tbp]
\caption{LLM Model Comparison}
\begin{center}
\small

\begin{NiceTabular}{Ip{1.7cm}Ip{1.1cm}Ip{0.4cm}Ip{1.1cm}Ip{1.2cm}Ip{0.95cm}I}
\hline

\cellcolor{headerblue}\textcolor{headerwhite}{Model}&Developer&Ref.&Release&Parameters&Context (tokens)\\
\Xhline{\ThickXhline pt}

\cellcolor{headerblue}\textcolor{headerwhite}{Claude Sonnet-4.5}&Anthropic&\cite{claude-sonnet4.5}&Sep-2025&``frontier''&200K\\
\hline
\cellcolor{headerblue}\textcolor{headerwhite}{Gemini2.5-Pro}&Google&\cite{gemini2.5}&Mar-2025&``large''&1M\\
\hline
\cellcolor{headerblue}\textcolor{headerwhite}{Gemini2.5-Flash}&Google&\cite{gemini2.5}&Jun-2025&``small'' est.17B&1M\\
\hline
\cellcolor{headerblue}\textcolor{headerwhite}{ChatGPT-5}&OpenAI&\cite{chatgpt5}&Aug-2025&``frontier''&400K\\
\hline
\cellcolor{headerblue}\textcolor{headerwhite}{ChatGPT-5-mini}&OpenAI&\cite{chatgpt5}&Aug-2025&``small''&400K\\
\hline
\cellcolor{headerblue}\textcolor{headerwhite}{Llama-4-maverick17b-128e-instruct}&Meta&\cite{llama4maverickhf}&Apr-2025&17B active, 400B total&128K\\
\hline
\cellcolor{headerblue}\textcolor{headerwhite}{Llama-3-405b-instruct}&Meta&\cite{grattafiori2024llama3herdmodels}&Jul-2024&405B&128K\\
\hline

\end{NiceTabular}
\label{tab:models}
\end{center}
\end{table}

%% file: fig_rtr.tex

\begin{figure*}[tp]
\centerline{\includegraphics[scale=0.5]{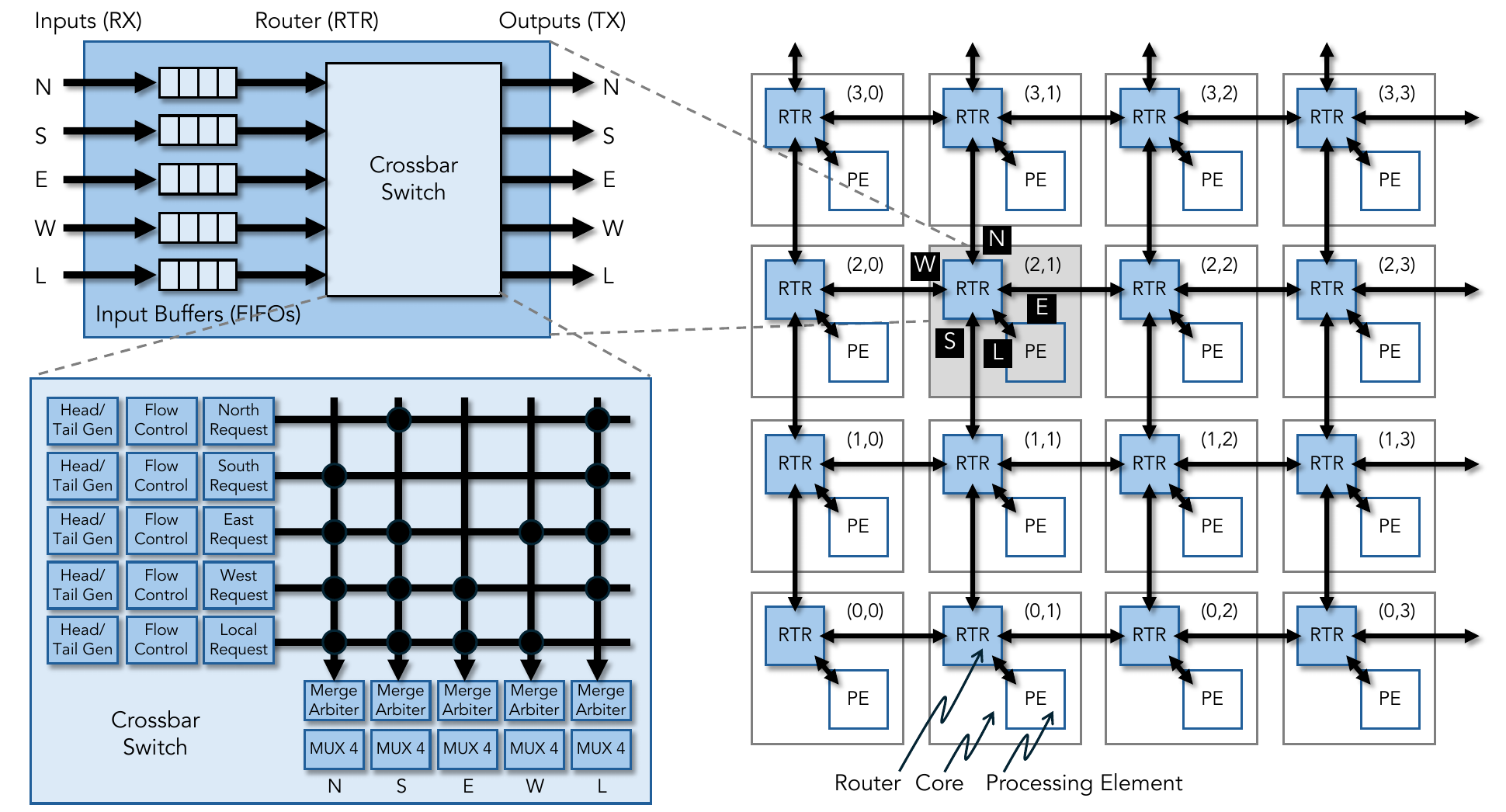}}
\caption{2D NoC Router Design: (Right) Two dimensions array of cores,
each core contains a router (RTR) and a processing element (PE).
Each core connects to neighboring cores via North (N) and South (S) connections in the Y-dimension,
East (E) and West (W) connections in the X-dimension, as well as from the core router to the 
core processing element via the Local (L) connection.  (Upper Left) The Router is comprised of 
five input channels with input buffers and five output channels, interconnected via a crossbar switch.
(Lower Left) The crossbar switch contains the logic for routing the packets.
}
\label{fig:2D_router}
\end{figure*}

%% file: experiment1_RTR.tex

\section{Experiment: 2D Network-on-Chip Router} \label{sec:experiments}

We tested our approach on a 2D Mesh Network-on-Chip (NoC) Router design,
comprised of 13 hierarchical design units,
and representative in complexity of an industry design.
%
The router design was specified via LCTs by a human logic designer.
The LCTs were given to an LLM to generate Verilog HDL in the forward transform.
The generated Verilog HDL was then given to an LLM to regenerate the original LCT specification in the inverse transform.

The 2D NoC Router design is depicted in Fig. \ref{fig:2D_router}.
In the context of a 2D core array, adjacent cores in the North, South, East, and West
directions are interconnected via a 2D router (RTR).  The router also has a Local 
connection from the 2D network to a local Processing Element (PE) in the same core.
All links are bidirectional and the TX link from one core connects to the RX link of
the adjacent core.
The router is composed of five input FIFOs and a non-blocking crossbar switch.
We divided the router into eleven logic design units
($FIFO$, $reg\_stage$, $flow\_control$, $head\_tail\_gen$,
$mrg\_arb$, $mux4$, $request\_N$,  $request\_S$,  $request\_E$,  $request\_W$,  $request\_L$), 
and two hierarchical connectivity units ($request\_top$, $rtr\_top$).
The function of each logic design unit is specified by an LCT.
The specification of each connectivity unit, which contains no logic, 
is a mapping table from source unit and signal to destination unit and signal.

The 2D NoC is designed to support variable length packets and a dimension-ordered
X-Y routing protocol, which routes horizontally, followed by vertically, followed by locally.
The first word of the packet is the header word, comprised of three fields:
the X-address for East-West routing, the Y-address for North-South routing,
and the length field which specifies the packet payload size in words.
The FIFO is a standard synchronous, dual-port FIFO.
Using the length field, the $head\_tail\_gen$ unit generates
a single-cycle ``head'' signal that aligns with the first word in the packet, and 
a single-cycle ``tail'' signal that aligns with the last word in the packet.
The request units perform X-Y address comparisons with the core X-Y IDs
to implement the routing rules and generate requests to the arbiter.
The $mrg\_arb$ unit arbitrates between incoming requests to a given output channel
and issues a grant based on a round-robin policy.  Arbitration is only performed
on the head word of packets, and maintained until the tail word is received.
Based on the selected incoming request, the $mux4$ unit outputs the correct packet.
In conjunction with the $reg\_stage$ unit, the $flow\_control$ unit meters
the flow of input packets into the crossbar based on the grant signals.
All units use $valid$ and $ready$ signaling for packet flow control within
the router unit and on \{N,S,E,W,L\} routing links.

%% file: results.tex

\vspace{-6pt}
\section{Results} \label{sec:results}

The results for LLM logic generation using seven different models are summarized in
Table \ref{tab:rtr_results}, using functional simulation to check the logic. 
There are 11 logic units and 2 connectivity units. 
%
Using only the Pass@1 metric \cite{chen2021evaluating} for both syntax and functionality,
two LLM models correctly designed all 13 units with 100\% success. 
Two LLM models had only single compile errors
(requiring only 2 lines of code each to be fixed). 
Two additional LLM models had insufficient context length to generate the largest design unit. 
Aside from the insufficient context length, all units were fully functional with less than 15 lines of code
modified (out of 1500-2000 lines of code generated, depending on the LLM model).

\input{table_rtr_results.tex}

\input{table_rtr_inv_results.tex}

The inverse transform from HDL to LCT was performed by the Gemini2.5-Pro model for all units,
across all HDL generating (forward transform) models.
Using different LLMs in the forward and inverse transforms separates the design and the check, 
similar to separate human designers and verifiers in traditional logic design and verification methodology.
The $reg\_stg$ and $fifo$ units were split
into multiple LCTs for recreation by process statement, matching the generation LCT structure.
The inverse transform results are summarized in Table \ref{tab:rtr_inv_results},
demonstrating not only the ability to reconstruct the LCT specification from the LLM
generated HDL logic, but also the ability to detect errors of hallucination and omission.
Both types of error, hallucination and omission, can occur in the
forward transform, inverse transform, and/or designer specification.
An example hallucination in the forward transform:
invalid routing rules (model Llama-3, unit $req\_E$).
An example omission in the forward transform:
priority encoder cases dropped during HDL generation (model Llama-4 unit $mrg\_arb$) .

Comparing the original and reconstructed LCTs in Table \ref{tab:rtr_inv_results},
with respect to Table \ref{tab:rtr_results},
the following cases emerge:
\begin{enumerate}
\item MATCH: M = 100\% match between original LCT and reconstructed LCT,
	indicating correctly generated logic.
    \begin{itemize}
        \item Green = Forward transform correct, inverse transform correct, and designer specification correct.
    \end{itemize}

\item MATCH: M SP = 100\% match between original LCT and reconstructed LCT,
	however, functional simulation indicated a failure (Table \ref{tab:rtr_results}).
	Closer inspection revealed subtle errors in the specification, not defined by the LCT (for example, a bitwidth mismatch).
	If the original LCT does not contain the information, the HDL cannot be reliably generated,
	nor the LCT reliably reconstructed.
    \begin{itemize}
        \item Yellow = Designer specification error.
    \end{itemize}

\item MISMATCH: X EQ = mismatching comparison between LCTs, however,
	closer inspection revealed logically and functionally equivalent tables.
	This particular case was due to an unused output specified in the original LCT.
	During HDL generation, the unused output was dropped from the HDL,
	and thus not included in the reconstructed LCT.
    \begin{itemize}
        \item Yellow = Designer specification error.
    \end{itemize}

\item MISMATCH: X FW = mismatching comparison between LCTs 
	correctly caught the functional error in the forward transform generating the HDL.
    \begin{itemize}
        \item Orange = Forward transform error.
    \end{itemize}

\item MISMATCH: X FW$\sim$S = mismatching comparison between LCTs 
	correctly caught errors not caught in functional simulation, due to missing testbench test cases.
    \begin{itemize}
        \item Orange = Forward transform error.
    \end{itemize}

\item MISMATCH: X INV = mismatching comparison between LCTs
	due to INV transform error.  The specific case found corresponds to a forward transform syntax error,
	which may have hindered the performance of the inverse transform.
    \begin{itemize}
        \item Blue = Inverse transform error.
    \end{itemize}
\end{enumerate}
Note that when comparing Table \ref{tab:rtr_inv_results} versus Table \ref{tab:rtr_results},
the focus is on correctly identifying the functional errors (F),
ignoring the syntax errors (S) in the forward transform.
In addition, the units $req\_top$ and $rtr\_top$ are connectivity tables, as opposed to LCTs,
and were not checked,
but there is nothing fundamentally blocking their evaluation with the same flow.

The following additional results did not affect the MATCH / MISMATCH comparison and were not annotated in
Table \ref{tab:rtr_inv_results}:
\begin{itemize}
\item Reordering (permuting) the rows and/or columns of an LCT is a valid table transformation that 
	does not change the functional specification of the LCT.

\item Another valid LCT transform is to expand ``don't care'' cases into enumerated cases.
	This changes the LCT contents, but generates a functionally equivalent LCT.

\item Minor specification gaps in the original LCTs were identified (e.g. missing reset case),
	but did not affect the results.

\item Minor differences in signal names, such as abbreviations or suffixes were inconsequential and ignored.
	Similarly, differences in numerical types, such as $3'd5$ and $3'b101$,
	that are logically equivalent are also ignored.

\item Additional default cases for fully specified case statements were inserted by the LLMs to cover
	uninitialized signals ($X's$), but have no bearing on functional logic.

\end{itemize}
While these issues prevent direct one-to-one comparison of LCTs, we do not see any significant
technical hurdles blocking automation of the LCT comparison step to determine logical equivalence,
completing the closed-loop methodology (Section \ref{subsec:close_loop}).

Note that it is possible that an error in the forward transform could be cancelled out by the
opposite error in the inverse transform, leading to an undetected forward transform error.
While unlikely, using different LLM models for the forward and inverse transforms
(as done in Table \ref{tab:rtr_inv_results})
reduces this possibility even further.

%% file: table_rtr_results.tex

\newcommand{\passcolor}{green!40}

\begin{table*}[tbp]
\caption{Router Experiment Results: Syntax errors in generated logic are denoted as S.
Functional errors in generated logic are denoted as F.
The number of errors is denoted as \#E and the number of lines of code to fix the errors are denoted as \#L. DNF is Did Not Finish, due to insufficient context length.}
\small
\begin{center}

\begin{NiceTabular}{IlIcIcIcIcIcIcIcIcIcIcIcIcIcI}
\hline

\cellcolor{headerblue}\textcolor{headerwhite}{Forward Model}&reg\_stg&flow\_ctrl&head\_tail&mrg\_arb&mux4&req\_N&req\_S&req\_E&req\_W&req\_L&fifo&req\_top&rtr\_top\\
\Xhline{\ThickXhline pt}

\cellcolor{headerblue}\textcolor{headerwhite}{ClaudeSonnet-4.5}&\cellcolor{\passcolor}{PASS}&\cellcolor{\passcolor}{PASS}&\cellcolor{\passcolor}{PASS}&\cellcolor{\passcolor}{PASS}&\cellcolor{\passcolor}{PASS}&\cellcolor{\passcolor}{PASS}&\cellcolor{\passcolor}{PASS}&\cellcolor{\passcolor}{PASS}&\cellcolor{\passcolor}{PASS}&\cellcolor{\passcolor}{PASS}&\cellcolor{\passcolor}{PASS}&\cellcolor{\passcolor}{PASS}&\cellcolor{\passcolor}{PASS}\\
\hline
\cellcolor{headerblue}\textcolor{headerwhite}{Gemini2.5-Pro}&\cellcolor{\passcolor}{PASS}&\cellcolor{\passcolor}{PASS}&\cellcolor{\passcolor}{PASS}&\cellcolor{\passcolor}{PASS}&\cellcolor{\passcolor}{PASS}&\cellcolor{\passcolor}{PASS}&\cellcolor{\passcolor}{PASS}&\cellcolor{\passcolor}{PASS}&\cellcolor{\passcolor}{PASS}&\cellcolor{\passcolor}{PASS}&\cellcolor{\passcolor}{PASS}&\cellcolor{\passcolor}{PASS}&\cellcolor{\passcolor}{PASS}\\
\hline
\cellcolor{headerblue}\textcolor{headerwhite}{Gemini2.5-Flash}&\cellcolor{\passcolor}{PASS}&\cellcolor{\passcolor}{PASS}&\cellcolor{\passcolor}{PASS}&\cellcolor{\passcolor}{PASS}&\cellcolor{\passcolor}{PASS}&\cellcolor{\passcolor}{PASS}&\cellcolor{\passcolor}{PASS}&\cellcolor{\passcolor}{PASS}&\cellcolor{\passcolor}{PASS}&S:1E,2L&\cellcolor{\passcolor}{PASS}&\cellcolor{\passcolor}{PASS}&\cellcolor{\passcolor}{PASS}\\
\hline
\cellcolor{headerblue}\textcolor{headerwhite}{ChatGPT-5}&\cellcolor{\passcolor}{PASS}&\cellcolor{\passcolor}{PASS}&\cellcolor{\passcolor}{PASS}&\cellcolor{\passcolor}{PASS}&\cellcolor{\passcolor}{PASS}&\cellcolor{\passcolor}{PASS}&\cellcolor{\passcolor}{PASS}&\cellcolor{\passcolor}{PASS}&\cellcolor{\passcolor}{PASS}&S:1E,2L&\cellcolor{\passcolor}{PASS}&\cellcolor{\passcolor}{PASS}&\cellcolor{\passcolor}{PASS}\\
\hline
\cellcolor{headerblue}\textcolor{headerwhite}{ChatGPT-5-mini}&\cellcolor{\passcolor}{PASS}&\cellcolor{\passcolor}{PASS}&\makecell{S:2E,6L,\\F:1E,2L}&F:1E,1L&\cellcolor{\passcolor}{PASS}&\cellcolor{\passcolor}{PASS}&\cellcolor{\passcolor}{PASS}&\cellcolor{\passcolor}{PASS}&\cellcolor{\passcolor}{PASS}&\cellcolor{\passcolor}{PASS}&\makecell{S:1E,1L,\\F:3E,4L}&\cellcolor{\passcolor}{PASS}&\cellcolor{\passcolor}{PASS}\\
\hline
\cellcolor{headerblue}\textcolor{headerwhite}{Llama-4-maverick}&\cellcolor{\passcolor}{PASS}&\cellcolor{\passcolor}{PASS}&F:2E,2L&\cellcolor{\passcolor}{PASS}&\cellcolor{\passcolor}{PASS}&\cellcolor{\passcolor}{PASS}&\cellcolor{\passcolor}{PASS}&\cellcolor{\passcolor}{PASS}&\cellcolor{\passcolor}{PASS}&\cellcolor{\passcolor}{PASS}&F:3E,4L&\cellcolor{\passcolor}{PASS}&DNF\\
\hline
\cellcolor{headerblue}\textcolor{headerwhite}{Llama-3-405b}&\cellcolor{\passcolor}{PASS}&\cellcolor{\passcolor}{PASS}&\cellcolor{\passcolor}{PASS}&\cellcolor{\passcolor}{PASS}&\cellcolor{\passcolor}{PASS}&\cellcolor{\passcolor}{PASS}&\cellcolor{\passcolor}{PASS}&F:2E,2L&F:2E,2L&F:3E,3L&\cellcolor{\passcolor}{PASS}&\cellcolor{\passcolor}{PASS}&DNF\\
\hline

\end{NiceTabular}
\label{tab:rtr_results}
\end{center}
\end{table*}

%% file: table_rtr_inv_results.tex

\newcommand{\goodcolor}{orange!40}
\newcommand{\failcolor}{cyan!20}
\newcommand{\correctcolor}{orange!40}
\newcommand{\cautioncolor}{yellow!40}

\begin{table*}[tbp]
\caption{Comparison results between input LCT specification and reconstructed LCT specification
(generated by Gemini2.5-Pro LLM).
Table key:
green = all correct,
orange = forward transform error,
blue = inverse transform error,
yellow = designer specification error.
Notation:
M = correctly identified LCT match,
M SP = correctly identified LCT match---functional fail due to incomplete designer specification,
X EQ = correctly identified LCT mismatch but functionally equivalent,
X FW = correctly identified LCT mismatch in the forward transform,
X FW$\sim$S = correctly identified LCT mismatch in the forward transform that was not caught by functional simulation test cases, and
X INV = correctly identified LCT mismatch in the inverse transform.
}
\small
\begin{center}

\begin{NiceTabular}{IlI>{\centering\arraybackslash}p{0.25cm}I>{\centering\arraybackslash}p{0.25cm}I>{\centering\arraybackslash}p{0.9cm}I>{\centering\arraybackslash}p{0.95cm}I>{\centering\arraybackslash}p{1.0cm}I>{\centering\arraybackslash}p{0.60cm}I>{\centering\arraybackslash}p{0.60cm}I>{\centering\arraybackslash}p{0.70cm}I>{\centering\arraybackslash}p{0.75cm}I>{\centering\arraybackslash}p{0.75cm}I>{\centering\arraybackslash}p{0.75cm}I>{\centering\arraybackslash}p{0.60cm}I>{\centering\arraybackslash}p{0.60cm}I>{\centering\arraybackslash}p{0.50cm}I>{\centering\arraybackslash}p{1.0cm}I}
\hline

\cellcolor{headerblue}\textcolor{headerwhite}{Forward Model}&\multicolumn{2}{c|}{reg\_stg}&flow\_ctrl&head\_tail&mrg\_arb&mux4&req\_N&req\_S&req\_E&req\_W&req\_L&\multicolumn{4}{c|}{fifo} \\

\Xhline{\ThickXhline pt}

\cellcolor{headerblue}\textcolor{headerwhite}{Claude-Sonnet-4.5}&\cellcolor{\passcolor}{M}&\cellcolor{\passcolor}{M}&\cellcolor{\passcolor}{M}&\cellcolor{\passcolor}{M}&\cellcolor{\passcolor}{M}&\cellcolor{\passcolor}{M}&\cellcolor{\passcolor}{M}&\cellcolor{\passcolor}{M}&\cellcolor{\passcolor}{M}&\cellcolor{\passcolor}{M}&\cellcolor{\passcolor}{M}&\cellcolor{\passcolor}{M}&\cellcolor{\passcolor}{M}&\cellcolor{\passcolor}{M}&\cellcolor{\passcolor}{M} \\
\hline
\cellcolor{headerblue}\textcolor{headerwhite}{Gemini2.5-Pro}&\cellcolor{\passcolor}{M}&\cellcolor{\passcolor}{M}&\cellcolor{\passcolor}{M}&\cellcolor{\passcolor}{M}&\cellcolor{\passcolor}{M}&\cellcolor{\passcolor}{M}&\cellcolor{\passcolor}{M}&\cellcolor{\cautioncolor}{X EQ}&\cellcolor{\passcolor}{M}&\cellcolor{\passcolor}{M}&\cellcolor{\passcolor}{M}&\cellcolor{\passcolor}{M}&\cellcolor{\passcolor}{M}&\cellcolor{\passcolor}{M}&\cellcolor{\passcolor}{M} \\
\hline
\cellcolor{headerblue}\textcolor{headerwhite}{Gemini2.5-Flash}&\cellcolor{\passcolor}{M}&\cellcolor{\passcolor}{M}&\cellcolor{\passcolor}{M}&\cellcolor{\passcolor}{M}&\cellcolor{\passcolor}{M}&\cellcolor{\passcolor}{M}&\cellcolor{\passcolor}{M}&\cellcolor{\cautioncolor}{X EQ}&\cellcolor{\passcolor}{M}&\cellcolor{\passcolor}{M}&\cellcolor{\passcolor}{M}&\cellcolor{\passcolor}{M}&\cellcolor{\passcolor}{M}&\cellcolor{\passcolor}{M}&\cellcolor{\correctcolor}{X FW$\sim$S} \\
\hline
\cellcolor{headerblue}\textcolor{headerwhite}{ChatGPT-5}&\cellcolor{\passcolor}{M}&\cellcolor{\passcolor}{M}&\cellcolor{\passcolor}{M}&\cellcolor{\passcolor}{M}&\cellcolor{\passcolor}{M}&\cellcolor{\passcolor}{M}&\cellcolor{\passcolor}{M}&\cellcolor{\passcolor}{M}&\cellcolor{\passcolor}{M}&\cellcolor{\passcolor}{M}&\cellcolor{\failcolor}{X INV}&\cellcolor{\passcolor}{M}&\cellcolor{\passcolor}{M}&\cellcolor{\passcolor}{M}&\cellcolor{\passcolor}{M} \\
\hline
\cellcolor{headerblue}\textcolor{headerwhite}{ChatGPT-5-mini}&\cellcolor{\passcolor}{M}&\cellcolor{\passcolor}{M}&\cellcolor{\passcolor}{M}&\cellcolor{\correctcolor}{X FW}&\cellcolor{\cautioncolor}{M SP}&\cellcolor{\passcolor}{M}&\cellcolor{\passcolor}{M}&\cellcolor{\cautioncolor}{X EQ}&\cellcolor{\passcolor}{M}&\cellcolor{\passcolor}{M}&\cellcolor{\passcolor}{M}&\cellcolor{\cautioncolor}{M SP}&\cellcolor{\cautioncolor}{M SP}&\cellcolor{\passcolor}{M}&\cellcolor{\passcolor}{M} \\
\hline
\cellcolor{headerblue}\textcolor{headerwhite}{Llama-4-maverick}&\cellcolor{\passcolor}{M}&\cellcolor{\passcolor}{M}&\cellcolor{\passcolor}{M}&\cellcolor{\correctcolor}{X FW}&\cellcolor{\goodcolor}{X FW$\sim$S}&\cellcolor{\passcolor}{M}&\cellcolor{\passcolor}{M}&\cellcolor{\cautioncolor}{X EQ}&\cellcolor{\passcolor}{M}&\cellcolor{\passcolor}{M}&\cellcolor{\passcolor}{M}&\cellcolor{\cautioncolor}{M SP}&\cellcolor{\cautioncolor}{M SP}&\cellcolor{\passcolor}{M}&\cellcolor{\passcolor}{M} \\
\hline
\cellcolor{headerblue}\textcolor{headerwhite}{Llama-3-405b}&\cellcolor{\passcolor}{M}&\cellcolor{\passcolor}{M}&\cellcolor{\passcolor}{M}&\cellcolor{\passcolor}{M}&\cellcolor{\goodcolor}{X FW$\sim$S}&\cellcolor{\passcolor}{M}&\cellcolor{\passcolor}{M}&\cellcolor{\passcolor}{M}&\cellcolor{\correctcolor}{X FW}&\cellcolor{\correctcolor}{X FW}&\cellcolor{\correctcolor}{X FW}&\cellcolor{\passcolor}{M}&\cellcolor{\passcolor}{M}&\cellcolor{\passcolor}{M}&\cellcolor{\passcolor}{M} \\
\hline

\end{NiceTabular}
\label{tab:rtr_inv_results}
\end{center}
\end{table*}

%% file: discussion.tex
\section{Discussion} \label{sec:discussion}

Summarizing the results, we applied an LLM forward transform to LCTs,
generating Verilog HDL to design a fully functional 2D NoC Router.
Then, after sufficiently augmenting the prompt to teach LLMs about LCTs by example,
we successfully used the LLM inverse transform to reconstruct the LCT specifications.
Comparing original and reconstructed LCTs successfully detected all five functional errors
in the forward transform found by simulation, while also detecting three more forward transform
errors, not found by simulation.
The reconstruction step only introduced one inverse transform error for additional review.
The final three functional errors were due to incomplete or ambiguous specification.
Our approach cannot be faulted for missing these, as it can only create and reconstruct based on 
the information that it has.  No penalty for information it does not have.


Using LCTs for specification addresses five challenges for generating logic HDL using LLMs.
\begin{enumerate}
\item {\bf Scale}: LCTs are inherently modular. They can be composed using connectivity tables.
Hierarchy demonstrated with the 2D NoC Router design.


\item {\bf Complexity}: 
LCTs concisely capture a large number of input conditions and output results.
Complex forward-backward flow control demonstrated with the 2D NoC Router design
and scaling complexity into tables with hundreds of cells.

\item {\bf Completeness}: 
An LCT with full coverage of all cases, conditions, and results is a completely specified design.

\item {\bf Specificity}: 
LCT cases (table rows) precisely define logical relationships from input conditions to output results,
leaving no room for ambiguous interpretation of function.

\item {\bf Reproducibility}: 
By precisely defining logical function,
LCTs are a reproducible specification, invariant across models, runs, and design styles.

\end{enumerate}


We learned two general lessons for generating useable logic with LLMs.
First, it is crucial for designers to be complete and precise in specification.
Anything left unspecified or ambiguously specified is open to
misinterpretation and misimplementation.
%
Second, breaking down designs into sufficiently small levels of complexity
contributed to successful LLM logic generation.
The larger the complexity of the design unit, the higher the chance of generation mistakes.
In our experimental 2D NoC Router design,
the size of the LCTs generally comprised tens of table cells (rows $\times$ columns),
ranging up to several hundred table cells for the merge arbiter.
Separate experiments demonstrated successful logic generation
(forward transform, checked by functional simulation)
of synthetic FSM LCTs into the range of thousands of table cells.

%% file: conclusion.tex
\section{Conclusion} \label{sec:conclusion}

The primary impact of this work is the demonstration of an invertible design flow using LLMs.
By encoding a specification from a source context into a destination context using an LLM,
and then decoding from the destination context back into the source context, also using an LLM,
the reconstructed specification can be directly compared with the original specification
to assert the veracity of the results.
We used this approach to successfully detect both matching and mismatching LCTs,
corresponding to functionally correct and functionally erroneous HDL respectively.
This development
represents an important catalyst for integrating LLMs in EDA design flows
to accelerate the development and verification time of logic hardware designs.
The forward (encoder) pass adds a new tool in accelerating design.
The inverse (decoder) pass adds a new tool in aiding correctness.
Together, they form a closed-loop design flow with built-in checking for errors due to LLM hallucinations and omissions.
In addition, the LCT specification itself is an important contribution to 
high-quality automated logic generation through complete, concise, and correct specification.


We demonstrated our results using 100\% publicly available, off-the-shelf LLMs.
We note that the top three performing models: Claude Sonnet4.5, Gemini2.5-Pro, and ChatGPT5
are models that use multi-step ``thinking'' and perform well on coding and complex reasoning tasks.
Going forward, we expect that 
LLM approaches that combine logic compilation, synthesis engines, simulations, and boolean logic
reasoning with pure LLM will outperform in logic generation.
Moreover, LLMs can be co-trained and fine-tuning to improve task performance on logic generation
using LCT specifications for further task performance gains.


While the large, frontier LLMs performed the best, we note the trend towards medium/small models,
for example, the 17B parameter Gemini2.5-Flash and the Llama4-maverick mixture-of-experts
with 17B active parameters performed reasonably well for a significantly smaller computational footprint.
We expect to see in the future more smaller models
\cite{belcak2025smalllanguagemodelsfuture,pareja2024unveilingsecretrecipeguide}
trained for domain specific tasks for optimized computational performance
\cite{wsg-2025-small-models}.
This bodes well for sustaining the energy costs of computation \cite{Noffsinger2025,Pilz2025,np_hpec_llm}
brought about by the demand for the latest AI capabilities.

%% file: references.bib
@article{np_science,
  title={Neural inference at the frontier of energy, space, and time},
  author={Modha, Dharmendra S and Akopyan, Filipp and Andreopoulos, Alexander and Appuswamy, Rathinakumar and Arthur, John V and Cassidy, Andrew S and Datta, Pallab and DeBole, Michael V and Esser, Steven K and Otero, Carlos Ortega and others},
  journal={Science},
  volume={382},
  number={6668},
  pages={329--335},
  year={2023},
  publisher={American Association for the Advancement of Science}
}

@inproceedings{np_isscc,
  title={{IBM NorthPole}: An Architecture for Neural Network Inference with a 12nm Chip},
  author={Cassidy, Andrew S and Arthur, John V and Akopyan, Filipp and Andreopoulos, Alexander and Appuswamy, Rathinakumar and Datta, Pallab and Debole, Michael V and Esser, Steven K and Otero, Carlos Ortega and Sawada, Jun and others},
  booktitle={2024 IEEE International Solid-State Circuits Conference (ISSCC)},
  volume={67},
  pages={214--215},
  year={2024},
  organization={IEEE}
}

@inproceedings{np_hpec_llm,
  author={Appuswamy, Rathinakumar and Debole, Michael V. and Taba, Brian and Esser, Steven K. and Cassidy, Andrew S. and Amir, Arnon and Andreopoulos, Alexander and Bablani, Deepika and Datta, Pallab and Kusnitz, Jeffrey A. and others},
  booktitle={2024 IEEE High Performance Extreme Computing Conference (HPEC)}, 
  title={Breakthrough Low-Latency, High-Energy-Efficiency {LLM} Inference Performance Using {NorthPole}}, 
  year={2024},
  volume={},
  number={},
  pages={1-8},
  doi={10.1109/HPEC62836.2024.10938418}
}

@online{gemini2.5,
  title     = {{Google Gemini}},
  author    = {{Google, Inc.}},
  year      = 2025,
  url       = {https://gemini.google.com},
  urldate   = {2025-10-30}
}

@online{claude-sonnet4.5,
  title     = {{Claude}},
  author    = {{Anthropic, PBC}},
  year      = 2025,
  url       = {https://claude.ai/},
  urldate   = {2025-10-30}
}

@online{chatgpt5,
  title     = {{GPT-5 is here}},
  author    = {{OpenAI}},
  year      = 2025,
  url       = {https://openai.com/gpt-5/},
  urldate   = {2025-11-08}
}

@misc{llama4maverickhf,
    title = {{meta-llama/Llama-4-Maverick-17B-128E-Original}},
    author = {{Meta AI}},
    howpublished = {\\url{
https://huggingface.co/meta-llama/Llama-4-Maverick-17B-128E-Original
}},
    year = {2025},
    month = apr,
    note = {Accessed: 2024-11-07}
}

@misc{grattafiori2024llama3herdmodels,
      title={{The Llama 3 Herd of Models}}, 
      author={Aaron Grattafiori and Abhimanyu Dubey and Abhinav Jauhri and Abhinav Pandey and Abhishek Kadian and Ahmad Al-Dahle and Aiesha Letman and Akhil Mathur and Alan Schelten and Alex Vaughan and Amy Yang and Angela Fan and Anirudh Goyal and Anthony Hartshorn and Aobo Yang and Archi Mitra and Archie Sravankumar and Artem Korenev and Arthur Hinsvark and Arun Rao and Aston Zhang and Aurelien Rodriguez and Austen Gregerson and Ava Spataru and Baptiste Roziere and Bethany Biron and Binh Tang and Bobbie Chern and Charlotte Caucheteux and Chaya Nayak and Chloe Bi and Chris Marra and Chris McConnell and Christian Keller and Christophe Touret and Chunyang Wu and Corinne Wong and Cristian Canton Ferrer and Cyrus Nikolaidis and Damien Allonsius and Daniel Song and Danielle Pintz and Danny Livshits and Danny Wyatt and David Esiobu and Dhruv Choudhary and Dhruv Mahajan and Diego Garcia-Olano and Diego Perino and Dieuwke Hupkes and Egor Lakomkin and Ehab AlBadawy and Elina Lobanova and Emily Dinan and Eric Michael Smith and Filip Radenovic and Francisco Guzmán and Frank Zhang and Gabriel Synnaeve and Gabrielle Lee and Georgia Lewis Anderson and Govind Thattai and Graeme Nail and Gregoire Mialon and Guan Pang and Guillem Cucurell and Hailey Nguyen and Hannah Korevaar and Hu Xu and Hugo Touvron and Iliyan Zarov and Imanol Arrieta Ibarra and Isabel Kloumann and Ishan Misra and Ivan Evtimov and Jack Zhang and Jade Copet and Jaewon Lee and Jan Geffert and Jana Vranes and Jason Park and Jay Mahadeokar and Jeet Shah and Jelmer van der Linde and Jennifer Billock and Jenny Hong and Jenya Lee and Jeremy Fu and Jianfeng Chi and Jianyu Huang and Jiawen Liu and Jie Wang and Jiecao Yu and Joanna Bitton and Joe Spisak and Jongsoo Park and Joseph Rocca and Joshua Johnstun and Joshua Saxe and Junteng Jia and Kalyan Vasuden Alwala and Karthik Prasad and Kartikeya Upasani and Kate Plawiak and Ke Li and Kenneth Heafield and Kevin Stone and Khalid El-Arini and Krithika Iyer and Kshitiz Malik and Kuenley Chiu and Kunal Bhalla and Kushal Lakhotia and Lauren Rantala-Yeary and Laurens van der Maaten and Lawrence Chen and Liang Tan and Liz Jenkins and Louis Martin and Lovish Madaan and Lubo Malo and Lukas Blecher and Lukas Landzaat and Luke de Oliveira and Madeline Muzzi and Mahesh Pasupuleti and Mannat Singh and Manohar Paluri and Marcin Kardas and Maria Tsimpoukelli and Mathew Oldham and Mathieu Rita and Maya Pavlova and Melanie Kambadur and Mike Lewis and Min Si and Mitesh Kumar Singh and Mona Hassan and Naman Goyal and Narjes Torabi and Nikolay Bashlykov and Nikolay Bogoychev and Niladri Chatterji and Ning Zhang and Olivier Duchenne and Onur Çelebi and Patrick Alrassy and Pengchuan Zhang and Pengwei Li and Petar Vasic and Peter Weng and Prajjwal Bhargava and Pratik Dubal and Praveen Krishnan and Punit Singh Koura and Puxin Xu and Qing He and Qingxiao Dong and Ragavan Srinivasan and Raj Ganapathy and Ramon Calderer and Ricardo Silveira Cabral and Robert Stojnic and Roberta Raileanu and Rohan Maheswari and Rohit Girdhar and Rohit Patel and Romain Sauvestre and Ronnie Polidoro and Roshan Sumbaly and Ross Taylor and Ruan Silva and Rui Hou and Rui Wang and Saghar Hosseini and Sahana Chennabasappa and Sanjay Singh and Sean Bell and Seohyun Sonia Kim and Sergey Edunov and Shaoliang Nie and Sharan Narang and Sharath Raparthy and Sheng Shen and Shengye Wan and Shruti Bhosale and Shun Zhang and Simon Vandenhende and Soumya Batra and Spencer Whitman and Sten Sootla and Stephane Collot and Suchin Gururangan and Sydney Borodinsky and Tamar Herman and Tara Fowler and Tarek Sheasha and Thomas Georgiou and Thomas Scialom and Tobias Speckbacher and Todor Mihaylov and Tong Xiao and Ujjwal Karn and Vedanuj Goswami and Vibhor Gupta and Vignesh Ramanathan and Viktor Kerkez and Vincent Gonguet and Virginie Do and Vish Vogeti and Vítor Albiero and Vladan Petrovic and Weiwei Chu and Wenhan Xiong and Wenyin Fu and Whitney Meers and Xavier Martinet and Xiaodong Wang and Xiaofang Wang and Xiaoqing Ellen Tan and Xide Xia and Xinfeng Xie and Xuchao Jia and Xuewei Wang and Yaelle Goldschlag and Yashesh Gaur and Yasmine Babaei and Yi Wen and Yiwen Song and Yuchen Zhang and Yue Li and Yuning Mao and Zacharie Delpierre Coudert and Zheng Yan and Zhengxing Chen and Zoe Papakipos and Aaditya Singh and Aayushi Srivastava and Abha Jain and Adam Kelsey and Adam Shajnfeld and Adithya Gangidi and Adolfo Victoria and Ahuva Goldstand and Ajay Menon and Ajay Sharma and Alex Boesenberg and Alexei Baevski and Allie Feinstein and Amanda Kallet and Amit Sangani and Amos Teo and Anam Yunus and Andrei Lupu and Andres Alvarado and Andrew Caples and Andrew Gu and Andrew Ho and Andrew Poulton and Andrew Ryan and Ankit Ramchandani and Annie Dong and Annie Franco and Anuj Goyal and Aparajita Saraf and Arkabandhu Chowdhury and Ashley Gabriel and Ashwin Bharambe and Assaf Eisenman and Azadeh Yazdan and Beau James and Ben Maurer and Benjamin Leonhardi and Bernie Huang and Beth Loyd and Beto De Paola and Bhargavi Paranjape and Bing Liu and Bo Wu and Boyu Ni and Braden Hancock and Bram Wasti and Brandon Spence and Brani Stojkovic and Brian Gamido and Britt Montalvo and Carl Parker and Carly Burton and Catalina Mejia and Ce Liu and Changhan Wang and Changkyu Kim and Chao Zhou and Chester Hu and Ching-Hsiang Chu and Chris Cai and Chris Tindal and Christoph Feichtenhofer and Cynthia Gao and Damon Civin and Dana Beaty and Daniel Kreymer and Daniel Li and David Adkins and David Xu and Davide Testuggine and Delia David and Devi Parikh and Diana Liskovich and Didem Foss and Dingkang Wang and Duc Le and Dustin Holland and Edward Dowling and Eissa Jamil and Elaine Montgomery and Eleonora Presani and Emily Hahn and Emily Wood and Eric-Tuan Le and Erik Brinkman and Esteban Arcaute and Evan Dunbar and Evan Smothers and Fei Sun and Felix Kreuk and Feng Tian and Filippos Kokkinos and Firat Ozgenel and Francesco Caggioni and Frank Kanayet and Frank Seide and Gabriela Medina Florez and Gabriella Schwarz and Gada Badeer and Georgia Swee and Gil Halpern and Grant Herman and Grigory Sizov and Guangyi and Zhang and Guna Lakshminarayanan and Hakan Inan and Hamid Shojanazeri and Han Zou and Hannah Wang and Hanwen Zha and Haroun Habeeb and Harrison Rudolph and Helen Suk and Henry Aspegren and Hunter Goldman and Hongyuan Zhan and Ibrahim Damlaj and Igor Molybog and Igor Tufanov and Ilias Leontiadis and Irina-Elena Veliche and Itai Gat and Jake Weissman and James Geboski and James Kohli and Janice Lam and Japhet Asher and Jean-Baptiste Gaya and Jeff Marcus and Jeff Tang and Jennifer Chan and Jenny Zhen and Jeremy Reizenstein and Jeremy Teboul and Jessica Zhong and Jian Jin and Jingyi Yang and Joe Cummings and Jon Carvill and Jon Shepard and Jonathan McPhie and Jonathan Torres and Josh Ginsburg and Junjie Wang and Kai Wu and Kam Hou U and Karan Saxena and Kartikay Khandelwal and Katayoun Zand and Kathy Matosich and Kaushik Veeraraghavan and Kelly Michelena and Keqian Li and Kiran Jagadeesh and Kun Huang and Kunal Chawla and Kyle Huang and Lailin Chen and Lakshya Garg and Lavender A and Leandro Silva and Lee Bell and Lei Zhang and Liangpeng Guo and Licheng Yu and Liron Moshkovich and Luca Wehrstedt and Madian Khabsa and Manav Avalani and Manish Bhatt and Martynas Mankus and Matan Hasson and Matthew Lennie and Matthias Reso and Maxim Groshev and Maxim Naumov and Maya Lathi and Meghan Keneally and Miao Liu and Michael L. Seltzer and Michal Valko and Michelle Restrepo and Mihir Patel and Mik Vyatskov and Mikayel Samvelyan and Mike Clark and Mike Macey and Mike Wang and Miquel Jubert Hermoso and Mo Metanat and Mohammad Rastegari and Munish Bansal and Nandhini Santhanam and Natascha Parks and Natasha White and Navyata Bawa and Nayan Singhal and Nick Egebo and Nicolas Usunier and Nikhil Mehta and Nikolay Pavlovich Laptev and Ning Dong and Norman Cheng and Oleg Chernoguz and Olivia Hart and Omkar Salpekar and Ozlem Kalinli and Parkin Kent and Parth Parekh and Paul Saab and Pavan Balaji and Pedro Rittner and Philip Bontrager and Pierre Roux and Piotr Dollar and Polina Zvyagina and Prashant Ratanchandani and Pritish Yuvraj and Qian Liang and Rachad Alao and Rachel Rodriguez and Rafi Ayub and Raghotham Murthy and Raghu Nayani and Rahul Mitra and Rangaprabhu Parthasarathy and Raymond Li and Rebekkah Hogan and Robin Battey and Rocky Wang and Russ Howes and Ruty Rinott and Sachin Mehta and Sachin Siby and Sai Jayesh Bondu and Samyak Datta and Sara Chugh and Sara Hunt and Sargun Dhillon and Sasha Sidorov and Satadru Pan and Saurabh Mahajan and Saurabh Verma and Seiji Yamamoto and Sharadh Ramaswamy and Shaun Lindsay and Shaun Lindsay and Sheng Feng and Shenghao Lin and Shengxin Cindy Zha and Shishir Patil and Shiva Shankar and Shuqiang Zhang and Shuqiang Zhang and Sinong Wang and Sneha Agarwal and Soji Sajuyigbe and Soumith Chintala and Stephanie Max and Stephen Chen and Steve Kehoe and Steve Satterfield and Sudarshan Govindaprasad and Sumit Gupta and Summer Deng and Sungmin Cho and Sunny Virk and Suraj Subramanian and Sy Choudhury and Sydney Goldman and Tal Remez and Tamar Glaser and Tamara Best and Thilo Koehler and Thomas Robinson and Tianhe Li and Tianjun Zhang and Tim Matthews and Timothy Chou and Tzook Shaked and Varun Vontimitta and Victoria Ajayi and Victoria Montanez and Vijai Mohan and Vinay Satish Kumar and Vishal Mangla and Vlad Ionescu and Vlad Poenaru and Vlad Tiberiu Mihailescu and Vladimir Ivanov and Wei Li and Wenchen Wang and Wenwen Jiang and Wes Bouaziz and Will Constable and Xiaocheng Tang and Xiaojian Wu and Xiaolan Wang and Xilun Wu and Xinbo Gao and Yaniv Kleinman and Yanjun Chen and Ye Hu and Ye Jia and Ye Qi and Yenda Li and Yilin Zhang and Ying Zhang and Yossi Adi and Youngjin Nam and Yu and Wang and Yu Zhao and Yuchen Hao and Yundi Qian and Yunlu Li and Yuzi He and Zach Rait and Zachary DeVito and Zef Rosnbrick and Zhaoduo Wen and Zhenyu Yang and Zhiwei Zhao and Zhiyu Ma},
      year={2024},
      eprint={2407.21783},
      archivePrefix={arXiv},
      primaryClass={cs.AI},
      url={https://arxiv.org/abs/2407.21783}, 
}

@techreport{pollack1963analysis,
  title={Analysis of the decision rules in decision tables},
  author={Pollack, Solomon L},
  year={1963}
}

@article{pooch1974translation,
  title={Translation of decision tables},
  author={Pooch, Udo W},
  journal={ACM Computing Surveys (CSUR)},
  volume={6},
  number={2},
  pages={125--151},
  year={1974},
  publisher={ACM New York, NY, USA}
}

@article{codasyl1982modern,
  title={A Modern Appraisal of Decision Tables},
  author={{CODDASYL}},
  journal={Report of the Decision Table Task Group},
  pages={230--232},
  year={1982}
}

@article{vanthienen1997decision,
  title={Decision tables: refining the concept and a proposed standard},
  author={Vanthienen, J and Dries, E},
  journal={Communications of the ACM},
  year={1997}
}

@article{post1921introduction,
  title={Introduction to a general theory of elementary propositions},
  author={Post, Emil L},
  journal={American journal of mathematics},
  volume={43},
  number={3},
  pages={163--185},
  year={1921},
  publisher={JSTOR}
}

@book{wittgenstein1922,
  author = {Wittgenstein, Ludwig},
  title = {Tractatus Logico-Philosophicus},
  publisher = {Project Gutenberg},
  year = {2010},
  translator = {C.K. Ogden},
  note = {Original work published 1922}
}

@article{hartree1946eniac,
  title={{The ENIAC, an electronic computing machine}},
  author={Hartree, Douglas Rayner},
  journal={Nature},
  volume={158},
  number={4015},
  pages={500--506},
  year={1946},
  publisher={Nature Publishing Group UK London}
}

@techreport{vonNeumann:1945:EDVAC,
  author = {von Neumann, John},
  title = {{First Draft of a Report on the EDVAC}},
  institution = {{Moore School of Electrical Engineering, University of Pennsylvania}},
  address = {{Philadelphia, PA, USA}},
  year = {1945},
  number = {{Contract No. W-670-ORD-4926}},
  month = {jun},
}

@article{hoare1969axiomatic,
  title={An axiomatic basis for computer programming},
  author={Hoare, Charles Antony Richard},
  journal={Communications of the ACM},
  volume={12},
  number={10},
  pages={576--580},
  year={1969},
  publisher={ACM New York, NY, USA}
}


%% file: references_AI_RTL.bib
@misc{roy2025veritasdeterministicverilogcode,
     title={{Veritas: Deterministic Verilog Code Synthesis from LLM-Generated Conjunctive Normal Form}},
     author={Prithwish Basu Roy and Akashdeep Saha and Manaar Alam and Johann Knechtel and Michail Maniatakos and Ozgur Sinanoglu and Ramesh Karri},
     year={2025},
     eprint={2506.00005},
     archivePrefix={arXiv},
     primaryClass={cs.AR},
     url={https://arxiv.org/abs/2506.00005},
}

@misc{calzada2025verilogdblargesthighestqualitydataset,
     title={{VerilogDB: The Largest, Highest-Quality Dataset with a Preprocessing Framework for LLM-based RTL Generation}},
     author={Paul E. Calzada and Zahin Ibnat and Tanvir Rahman and Kamal Kandula and Danyu Lu and Sujan Kumar Saha and Farimah Farahmandi and Mark Tehranipoor},
     year={2025},
     eprint={2507.13369},
     archivePrefix={arXiv},
     primaryClass={cs.AR},
     url={https://arxiv.org/abs/2507.13369},
}

@misc{thakur2023verigenlargelanguagemodel,
     title={{VeriGen: A Large Language Model for Verilog Code Generation}},
     author={Shailja Thakur and Baleegh Ahmad and Hammond Pearce and Benjamin Tan and Brendan Dolan-Gavitt and Ramesh Karri and Siddharth Garg},
     year={2023},
     eprint={2308.00708},
     archivePrefix={arXiv},
     primaryClass={cs.PL},
     url={https://arxiv.org/abs/2308.00708},
}

@misc{wei2025vericoderenhancingllmbasedrtl,
     title={{VeriCoder: Enhancing LLM-Based RTL Code Generation through Functional Correctness Validation}},
     author={Anjiang Wei and Huanmi Tan and Tarun Suresh and Daniel Mendoza and Thiago S. F. X. Teixeira and Ke Wang and Caroline Trippel and Alex Aiken},
     year={2025},
     eprint={2504.15659},
     archivePrefix={arXiv},
     primaryClass={cs.AR},
     url={https://arxiv.org/abs/2504.15659},
}

@INPROCEEDINGS{min2025improving,
 author={Min, Kyungjun and Park, Seonghyeon and Park, Hyeonwoo and Cho, Jinoh and Kang, Seokhyeong},
 booktitle={{2025 Design, Automation \& Test in Europe Conference (DATE)}},
 title={{Improving LLM-Based Verilog Code Generation with Data Augmentation and RL}},
 year={2025},
 volume={},
 number={},
 pages={1-7},
 doi={10.23919/DATE64628.2025.10992897}
}

@misc{gao2024autovcodersystematicframeworkautomated,
     title={{AutoVCoder: A Systematic Framework for Automated Verilog Code Generation using LLMs}},
     author={Mingzhe Gao and Jieru Zhao and Zhe Lin and Wenchao Ding and Xiaofeng Hou and Yu Feng and Chao Li and Minyi Guo},
     year={2024},
     eprint={2407.18333},
     archivePrefix={arXiv},
     primaryClass={cs.AR},
     url={https://arxiv.org/abs/2407.18333},
}

@misc{zhao2025codevempoweringllmshdl,
     title={{CodeV: Empowering LLMs with HDL Generation through Multi-Level Summarization}},
     author={Yang Zhao and Di Huang and Chongxiao Li and Pengwei Jin and Muxin Song and Yinan Xu and Ziyuan Nan and Mingju Gao and Tianyun Ma and Lei Qi and Yansong Pan and Zhenxing Zhang and Rui Zhang and Xishan Zhang and Zidong Du and Qi Guo and Xing Hu},
     year={2025},
     eprint={2407.10424},
     archivePrefix={arXiv},
     primaryClass={cs.PL},
     url={https://arxiv.org/abs/2407.10424},
}

@article{blocklove2025automaticallyimproving,
author = {Blocklove, Jason and Thakur, Shailja and Tan, Benjamin and Pearce, Hammond and Garg, Siddharth and Karri, Ramesh},
title = {{Automatically Improving LLM-based Verilog Generation using EDA Tool Feedback}},
year = {2025},
issue_date = {November 2025},
publisher = {Association for Computing Machinery},
address = {New York, NY, USA},
volume = {30},
number = {6},
issn = {1084-4309},
url = {https://doi.org/10.1145/3723876},
doi = {10.1145/3723876},
journal = {ACM Trans. Des. Autom. Electron. Syst.},
month = oct,
articleno = {100},
numpages = {26}
}

@misc{Roberts2025,
title={{Improving LLM Performance in Generating Verilog by Fine Tuning with a Translated Code Dataset}},
author={Brendan Roberts},
url={https://www2.eecs.berkeley.edu/Pubs/TechRpts/2025/EECS-2025-104.pdf},
publisher={Electrical Engineering and Computer Sciences University of California, Berkeley},
year={2025},
month = may
}

@article{chen2021evaluating,
  author       = {Mark Chen and
                  Jerry Tworek and
                  Heewoo Jun and
                  Qiming Yuan and
                  Henrique Pond{\'{e}} de Oliveira Pinto and
                  Jared Kaplan and
                  Harri Edwards and
                  Yuri Burda and
                  Nicholas Joseph and
                  Greg Brockman and
                  Alex Ray and
                  Raul Puri and
                  Gretchen Krueger and
                  Michael Petrov and
                  Heidy Khlaaf and
                  Girish Sastry and
                  Pamela Mishkin and
                  Brooke Chan and
                  Scott Gray and
                  Nick Ryder and
                  Mikhail Pavlov and
                  Alethea Power and
                  Lukasz Kaiser and
                  Mohammad Bavarian and
                  Clemens Winter and
                  Philippe Tillet and
                  Felipe Petroski Such and
                  Dave Cummings and
                  Matthias Plappert and
                  Fotios Chantzis and
                  Elizabeth Barnes and
                  Ariel Herbert{-}Voss and
                  William Hebgen Guss and
                  Alex Nichol and
                  Alex Paino and
                  Nikolas Tezak and
                  Jie Tang and
                  Igor Babuschkin and
                  Suchir Balaji and
                  Shantanu Jain and
                  William Saunders and
                  Christopher Hesse and
                  Andrew N. Carr and
                  Jan Leike and
                  Joshua Achiam and
                  Vedant Misra and
                  Evan Morikawa and
                  Alec Radford and
                  Matthew Knight and
                  Miles Brundage and
                  Mira Murati and
                  Katie Mayer and
                  Peter Welinder and
                  Bob McGrew and
                  Dario Amodei and
                  Sam McCandlish and
                  Ilya Sutskever and
                  Wojciech Zaremba},
  title        = {Evaluating Large Language Models Trained on Code},
  journal      = {CoRR},
  volume       = {abs/2107.03374},
  year         = {2021},
  url          = {https://arxiv.org/abs/2107.03374},
  eprinttype    = {arXiv},
  eprint       = {2107.03374},
  bibsource    = {dblp computer science bibliography, https://dblp.org}
}


%% file: references_autoenc.bib
@misc{nikolaou2025languagemodelsinjectiveinvertible,
      title={Language Models are Injective and Hence Invertible}, 
      author={Giorgos Nikolaou and Tommaso Mencattini and Donato Crisostomi and Andrea Santilli and Yannis Panagakis and Emanuele Rodolà},
      year={2025},
      eprint={2510.15511},
      archivePrefix={arXiv},
      primaryClass={cs.LG},
      url={https://arxiv.org/abs/2510.15511}, 
}

@article{hecht1995replicator,
  title={Replicator neural networks for universal optimal source coding},
  author={Hecht-Nielsen, Robert},
  journal={Science},
  volume={269},
  number={5232},
  pages={1860--1863},
  year={1995},
  publisher={American Association for the Advancement of Science}
}

@misc{xu2025hallucinationinevitableinnatelimitation,
      title={Hallucination is Inevitable: An Innate Limitation of Large Language Models}, 
      author={Ziwei Xu and Sanjay Jain and Mohan Kankanhalli},
      year={2025},
      eprint={2401.11817},
      archivePrefix={arXiv},
      primaryClass={cs.CL},
      url={https://arxiv.org/abs/2401.11817}, 
}

@misc{fu2025absencebenchlanguagemodelscant,
      title={AbsenceBench: Language Models Can't Tell What's Missing}, 
      author={Harvey Yiyun Fu and Aryan Shrivastava and Jared Moore and Peter West and Chenhao Tan and Ari Holtzman},
      year={2025},
      eprint={2506.11440},
      archivePrefix={arXiv},
      primaryClass={cs.CL},
      url={https://arxiv.org/abs/2506.11440}, 
}

@inproceedings{bender2020climbing,
  title={Climbing towards NLU: On meaning, form, and understanding in the age of data},
  author={Bender, Emily M and Koller, Alexander},
  booktitle={Proceedings of the 58th annual meeting of the association for computational linguistics},
  pages={5185--5198},
  year={2020}
}

@misc{belcak2025smalllanguagemodelsfuture,
      title={Small Language Models are the Future of Agentic {AI}}, 
      author={Peter Belcak and Greg Heinrich and Shizhe Diao and Yonggan Fu and Xin Dong and Saurav Muralidharan and Yingyan Celine Lin and Pavlo Molchanov},
      year={2025},
      eprint={2506.02153},
      archivePrefix={arXiv},
      primaryClass={cs.AI},
      url={https://arxiv.org/abs/2506.02153}, 
}

@misc{pareja2024unveilingsecretrecipeguide,
      title={Unveiling the Secret Recipe: A Guide For Supervised Fine-Tuning Small {LLMs}}, 
      author={Aldo Pareja and Nikhil Shivakumar Nayak and Hao Wang and Krishnateja Killamsetty and Shivchander Sudalairaj and Wenlong Zhao and Seungwook Han and Abhishek Bhandwaldar and Guangxuan Xu and Kai Xu and others},
      year={2024},
      eprint={2412.13337},
      archivePrefix={arXiv},
      primaryClass={cs.LG},
      url={https://arxiv.org/abs/2412.13337}, 
}

@article{wsg-2025-small-models,
  author  = {Mims, Christopher},
  title   = {Large Language Models Get All the Hype, but Small Models Do the Real Work},
  journal = {The Wall Street Journal},
  year    = {2025},
  month   = oct,
  day     = {31},
  url     = {https://www.wsj.com/tech/ai/large-language-models-get-all-the-hype-but-small-models-do-the-real-work-225d3145}
}

@article{Noffsinger2025,
  author       = {Noffsinger, Jesse and Patel, Mark and Sachdeva, Pankaj and Bhan, Arjita and Chang, Haley and Goodpaster, Maria},
  title        = {The cost of compute: A \$7 trillion race to scale data centers},
  journal      = {McKinsey \& Company Insights},
  year         = {2025},
  month        = {April},
  day          = {28},
  url          = {https://www.mckinsey.com/industries/technology-media-and-telecommunications/our-insights/the-cost-of-compute-a-7-trillion-dollar-race-to-scale-data-centers}
}

@techreport{Pilz2025,
  author       = {Pilz, Konstantin F. and Mahmood, Yusuf and Heim, Lennart},
  title        = {AI's Power Requirements Under Exponential Growth: Extrapolating AI Data Center Power Demand and Assessing Its Potential Impact on U.S. Competitiveness},
  institution  = {RAND Corporation},
  number       = {RR-A3572-1},
  year         = {2025},
  url          = {https://www.rand.org/content/dam/rand/pubs/research_reports/RRA3500/RRA3572-1/RAND_RRA3572-1.pdf#:~:text=Given%20recent%20training%20compute%20growth%2C,the%20power%20capacity%20needed%20for},
  doi          = {10.7249/RRA3572-1}
}
